\documentclass[11pt,letterpaper]{article}

\usepackage[utf8]{inputenc}
\usepackage[T1]{fontenc}
\usepackage{amsmath,amssymb,amsfonts}
\usepackage{graphicx}
\usepackage{booktabs}
\usepackage[table]{xcolor}
\usepackage{hyperref}
\usepackage{algorithm}
\usepackage{algorithmic}
\usepackage{tikz}
\usepackage{pgfplots}
\usepackage[margin=1in]{geometry}
\usepackage{natbib}
\usepackage{lmodern}
\usepackage{microtype}
\usepackage{float}
\usetikzlibrary{shapes.geometric, arrows.meta, positioning, fit, backgrounds, calc}
\pgfplotsset{compat=1.17}

\definecolor{primaryblue}{RGB}{31,119,180}
\definecolor{accentorange}{RGB}{255,127,14}
\definecolor{successgreen}{RGB}{44,160,44}
\definecolor{dangerred}{RGB}{214,39,40}
\definecolor{purpleaccent}{RGB}{148,103,189}
\definecolor{lightgray}{RGB}{240,240,240}
\definecolor{boxgreen}{RGB}{230,255,230}

\hypersetup{
    colorlinks=true,
    linkcolor=primaryblue,
    citecolor=primaryblue,
    urlcolor=primaryblue
}

\title{Beyond Uniform Sampling: Synergistic Active Learning and Input Denoising for Robust Neural Operators}

\author{
    Samrendra Roy$^{1,\ast}$, \quad Souvik Chakraborty$^{2}$, \quad Syed Bahauddin Alam$^{1,3}$
    \\[1.2em]
    \small $^{1}$Department of Nuclear, Plasma, and Radiological Engineering,\\
    \small University of Illinois Urbana-Champaign, Urbana, IL, USA\\[0.4em]
    \small $^{2}$Department of Applied Mechanics,\\
    \small Indian Institute of Technology Delhi, New Delhi, India\\[0.4em]
    \small $^{3}$National Center for Supercomputing Applications,\\
    \small University of Illinois Urbana-Champaign, Urbana, IL, USA\\[0.6em]
    \small $^{\ast}$Corresponding author: \texttt{roysam@illinois.edu}
}

\date{April 2026}

\begin{document}

\maketitle

\begin{abstract}
Neural operators have emerged as fast surrogate models for physics simulations, yet they remain acutely vulnerable to adversarial perturbations, a critical liability for safety-critical digital twin deployments. We present a synergistic defense that combines \emph{active learning}-based data generation with an \emph{input denoising} architecture. The active learning component adaptively probes model weaknesses using differential evolution attacks, then generates targeted training data at discovered vulnerability locations while an adaptive smooth-ratio safeguard preserves baseline accuracy. The input denoising component augments the operator architecture with a learnable bottleneck that filters adversarial noise while retaining physics-relevant features. On the viscous Burgers' equation benchmark, the combined approach achieves a \textbf{2.04\%} combined error (1.21\% baseline + 0.83\% robustness), representing an \textbf{87\% reduction} relative to standard training (15.42\% combined) and outperforming both active learning alone (3.42\%) and input denoising alone (5.22\%). More broadly, our results, combined with cross-architecture vulnerability analysis from prior work, suggest that optimal training data for neural operators is architecture-dependent: because different architectures concentrate sensitivity in distinct input subspaces, uniform sampling cannot adequately cover the vulnerability landscape of all models. These findings have potential implications for the deployment of neural operators in safety-critical energy systems including nuclear reactor monitoring.
\end{abstract}


\section{Introduction}

Neural operators have emerged as powerful surrogate models that learn mappings between infinite-dimensional function spaces, enabling rapid inference for partial differential equations (PDEs) without expensive re-simulation~\citep{lu2021learning,kovachki2023neural}. Their ability to generalize across input functions, rather than individual discretizations, has driven adoption in applications spanning fluid dynamics, structural mechanics, and nuclear engineering~\citep{kobayashi2024deeponet,kobayashi2024generalization,hossain2024virtual,kobayashi2025proxies}. In these domains, neural operators serve as the predictive backbone of digital twin frameworks, where real-time inference from sparse sensor data is essential for monitoring and decision-making.

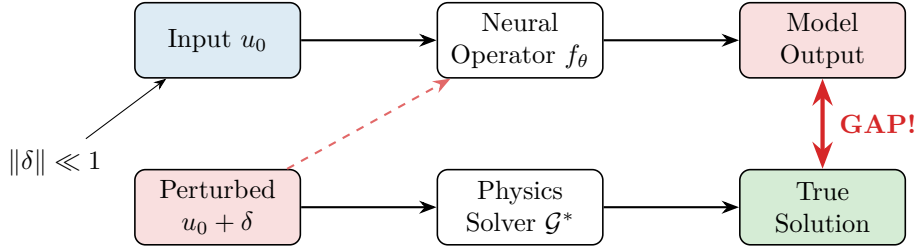
\begin{figure}[t]
\centering
\begin{tikzpicture}[
    node distance=0.8cm and 1.4cm,
    block/.style={rectangle, draw, rounded corners=4pt, minimum width=2.2cm, minimum height=1cm, align=center, font=\small, line width=0.6pt},
    arrow/.style={-Stealth, thick},
    dashedarrow/.style={-Stealth, thick, dashed, dangerred!70}
]
    \node[block, fill=primaryblue!15] (input) {Input $u_0$};
    
    \node[block, fill=dangerred!15, below=1.2cm of input] (perturbed) {Perturbed\\$u_0 + \delta$};
    
    \node[font=\small, left=0.3cm of perturbed, yshift=0.6cm] (delta) {$\|\delta\| \ll 1$};
    \draw[-Stealth] (delta) -- (input);
    
    \node[block, fill=white, right=1.8cm of input] (operator) {Neural\\Operator $f_\theta$};
    
    \node[block, fill=white, right=1.8cm of perturbed] (solver) {Physics\\Solver $\mathcal{G}^*$};
    
    \node[block, fill=dangerred!15, right=1.8cm of operator] (output) {Model\\Output};
    
    \node[block, fill=successgreen!20, right=1.8cm of solver] (true) {True\\Solution};
    
    \draw[arrow] (input) -- (operator);
    \draw[arrow] (operator) -- (output);
    \draw[dashedarrow] (perturbed) -- (operator);
    \draw[arrow] (perturbed) -- (solver);
    \draw[arrow] (solver) -- (true);
    
    \draw[dangerred, ultra thick, Stealth-Stealth] (output.south) -- node[right, font=\small\bfseries, dangerred, xshift=2pt] {GAP!} (true.north);
    
\end{tikzpicture}
\caption{The adversarial robustness problem: a small input perturbation $\delta$ causes the neural operator's prediction to diverge from the true physics solution, even though the well-posed PDE produces a nearly identical output for the perturbed input.}
\label{fig:problem}
\end{figure}

However, a recent study has exposed a critical vulnerability in this paradigm (Figure~\ref{fig:problem}): neural operators are acutely susceptible to adversarial perturbations~\citep{roy2026adversarial}. On a multi-physics CFD benchmark studied in that work, even models achieving relative $L_2$ errors below $10^{-5}$ on clean validation data suffered \emph{five orders of magnitude} performance degradation under sparse, physically plausible perturbations affecting fewer than 1\% of input dimensions. The fundamental concern is not merely that model accuracy degrades, but that the model \emph{deviates from the underlying physics}: when the same perturbed input is fed to both the neural operator and the physics solver, the solver produces a solution close to the unperturbed one (by continuous dependence on data), while the neural operator's prediction diverges catastrophically. This gap between model output and true physics (Figure~\ref{fig:problem}) grows not because the physics changed, but because the learned mapping amplifies perturbations that the PDE's continuous dependence property would bound. In that same study, 100\% of successful single-point attacks evaded standard anomaly detection, rendering them invisible to conventional monitoring pipelines. This poses an unacceptable risk for deployment in safety-critical energy systems, particularly nuclear thermal-hydraulic monitoring, where digital twins must deliver reliable predictions even under sensor noise, calibration drift, or deliberate adversarial manipulation.

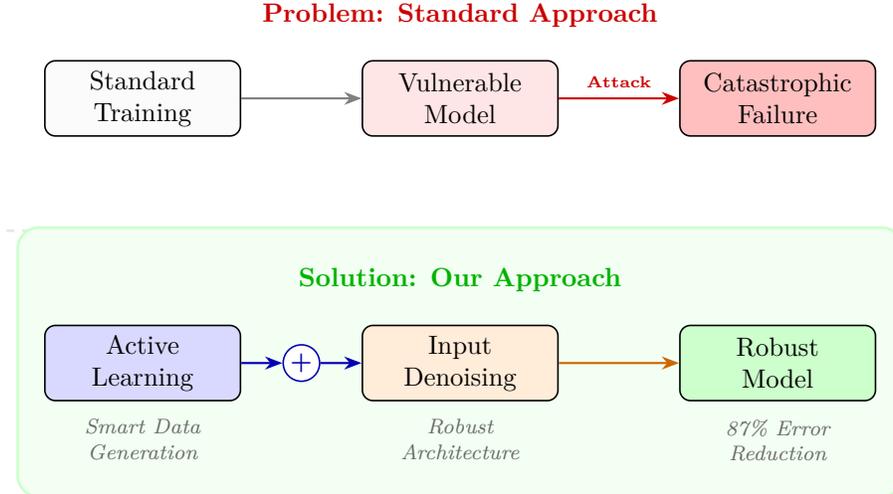
\begin{figure}[t]
\centering
\begin{tikzpicture}[
    node distance=1.1cm and 1.6cm,
    block/.style={
        rectangle, draw, rounded corners=4pt, 
        minimum width=2.6cm, minimum height=1.0cm, 
        align=center, font=\small, line width=0.6pt
    },
    arrow/.style={-Stealth, thick, line width=0.8pt},
    sublabel/.style={font=\scriptsize\itshape, color=gray!80!black, align=center},
    header/.style={font=\small\bfseries, inner sep=5pt},
    dangerred/.style={color=red!80!black},
    successgreen/.style={color=green!50!black},
    primaryblue/.style={color=blue!70!black},
    accentorange/.style={color=orange!80!black}
]

    \node[block, fill=lightgray!30] (standard) {Standard\\Training};
    \node[block, fill=red!10, right=of standard] (vuln) {Vulnerable\\Model};
    \node[block, fill=red!25, right=of vuln] (fail) {Catastrophic\\Failure};
    
    \node[header, red!80!black] (problemHeader) at ($(standard.north)!0.5!(fail.north) + (0,0.6)$) {Problem: Standard Approach};

    \draw[arrow, gray] (standard) -- (vuln);
    \draw[arrow, red!80!black] (vuln) -- node[above, font=\tiny\bfseries] {Attack} (fail);

    \node[block, fill=blue!15, below=2.5cm of standard] (al) {Active\\Learning};
    \node[block, fill=orange!15, right=of al] (denoise) {Input\\Denoising};
    \node[block, fill=green!20, right=of denoise] (robust) {Robust\\Model};

    \node[header, green!70!black] (solutionHeader) at ($(al.north)!0.5!(robust.north) + (0,0.6)$) {Solution: Our Approach};
    
    \path (al) -- node[circle, draw=blue!70!black, fill=white, inner sep=1pt, font=\footnotesize\bfseries, text=blue!70!black, line width=0.6pt] (plus) {+} (denoise);
    \draw[arrow, blue!70!black] (al) -- (plus);
    \draw[arrow, blue!70!black] (plus) -- (denoise);
    \draw[arrow, orange!80!black] (denoise) -- (robust);
    
    \node[sublabel, below=0.1cm of al] (descr1) {Smart Data\\Generation};
    \node[sublabel, below=0.1cm of denoise] (descr2) {Robust\\Architecture};
    \node[sublabel, below=0.1cm of robust] (descr3) {87\% Error\\Reduction};
    
    \begin{scope}[on background layer]
        \draw[gray!20, dashed, line width=1pt] 
            ($(standard.south west)!0.5!(al.north west) + (-0.5,0)$) -- 
            ($(fail.south east)!0.5!(robust.north east) + (0.5,0)$);

        \node[fill=green!5, draw=green!20, rounded corners=8pt, line width=1pt,
              fit=(solutionHeader) (al) (robust) (plus) (descr1) (descr2) (descr3), 
              inner sep=10pt] (box) {};
    \end{scope}
\end{tikzpicture}
\caption{Overview of our defense strategy. Standard training produces models that fail catastrophically under adversarial attack. Our approach combines active learning (smart data generation targeting discovered vulnerabilities) with input denoising (a learnable bottleneck providing architectural robustness) for synergistic defense.}
\label{fig:overview}
\end{figure}

The root cause of this vulnerability, as characterized by~\citet{roy2026adversarial}, is a \emph{sensitivity mismatch}: neural operators learn latent representations optimized for reconstruction fidelity rather than alignment with physics-relevant input distances. Consequently, perturbations that are small in physical terms can map to large displacements in the learned latent space. The \emph{effective perturbation dimension} $d_{\text{eff}}$, a Jacobian-derived diagnostic introduced in that work, further reveals that vulnerability is architecture-dependent. On their CFD benchmark, models with moderate sensitivity concentration (e.g., S-DeepONet, $d_{\text{eff}} \approx 4$) were more exploitable than those with extreme concentration (e.g., POD-DeepONet, $d_{\text{eff}} \approx 1$), because the latter's low-rank output projections inherently cap maximum error.

While~\citet{roy2026adversarial} established the attack surface, defense strategies for neural operators remain largely unexplored. In the broader deep learning literature, adversarial training~\citep{madry2018towards}, randomized smoothing~\citep{cohen2019certified}, and Lipschitz regularization~\citep{miyato2018spectral,gouk2021regularisation} have shown promise for classification networks, but their direct applicability to operator learning, where inputs and outputs are functions rather than vectors, is not straightforward.

In this paper, we present preliminary results on a synergistic defense combining two complementary strategies (Figure~\ref{fig:overview}). The first is \textbf{active learning for robust data generation}: rather than sampling training data uniformly, we iteratively probe model weaknesses using differential evolution (DE) attacks and generate targeted training data at discovered vulnerability locations, guided by an adaptive smooth-ratio safeguard that prevents baseline degradation. The second is an \textbf{input denoising architecture} that augments the DeepONet branch network with a learnable autoencoder bottleneck, compressing and reconstructing input functions to filter high-frequency adversarial perturbations while preserving physics-relevant features.

The key insight is that these mechanisms address complementary aspects of the vulnerability. Active learning teaches the model the correct perturbation-response mapping, specifically \emph{where} to be robust, by pairing perturbed inputs with their true physics solutions obtained from the simulator. Input denoising provides \emph{inherent} architectural robustness by reducing the effective dimensionality of the input representation, making the model less sensitive to perturbations irrespective of their location. Together, they yield stronger defense than either approach in isolation.

\section{Background}

\subsection{Neural Operators and DeepONet}

The Deep Operator Network (DeepONet)~\citep{lu2021learning} learns operators mapping between function spaces. For a PDE with initial condition $u_0 \in \mathcal{U}$, the true solution operator $\mathcal{G}^*[u_0](\mathbf{x}) = u(\mathbf{x}, T)$ maps to the solution at target time $T$. DeepONet approximates this operator via a learned model $f_\theta$:
\begin{equation}
    f_\theta[u_0](\mathbf{x}) = \sum_{k=1}^{p} b_k(u_0) \cdot t_k(\mathbf{x}) + b_0 \approx \mathcal{G}^*[u_0](\mathbf{x}),
\end{equation}
where $\{b_k\}_{k=1}^{p}$ are outputs of the \emph{branch network} encoding the input function at a fixed set of sensor locations, and $\{t_k\}_{k=1}^{p}$ are outputs of the \emph{trunk network} encoding spatial query coordinates. DeepONet and its variants have been successfully deployed for real-time inference in nuclear digital twin frameworks~\citep{kobayashi2024deeponet,kobayashi2024generalization}, virtual sensing~\citep{hossain2024virtual}, and cross-domain spatiotemporal forecasting~\citep{kobayashi2025proxies}. A comprehensive comparison of DeepONet variants is provided by~\citet{lu2022comprehensive}.

\subsection{Adversarial Vulnerability in Neural Operators}

\citet{roy2026adversarial} demonstrated that neural operators exhibit acute vulnerability to sparse adversarial perturbations discovered via gradient-free differential evolution (DE). Perturbations affecting fewer than 1\% of input dimensions can increase the relative $L_2$ error from approximately 1.5\% to 37--63\%, with attack success rates exceeding 98\% across four architectures (MIMONet, NOMAD, S-DeepONet, POD-DeepONet). Random perturbations of equal magnitude achieve near-zero success rates, confirming that the vulnerabilities are structural rather than a consequence of general sensitivity.

The effective perturbation dimension $d_{\text{eff}}$, defined via the decay rate of the Jacobian singular value spectrum, provides a compact diagnostic for differential vulnerability across architectures. Models with moderate $d_{\text{eff}}$ and high sensitivity concentration are most exploitable, while models with very low $d_{\text{eff}}$ exhibit capped error because their output is confined to a low-rank subspace (e.g., the POD basis in POD-DeepONet), limiting the damage any input perturbation can inflict.

Prior work on adversarial robustness in the neural operator setting is limited. \citet{adesoji2022evaluating} evaluated adversarial robustness of Fourier neural operators (FNOs), while \citet{zhu2023fourier} proposed Fourier-enhanced architectures with improved robustness properties. However, neither work provides a systematic defense framework applicable across operator architectures.

\section{Methodology}

\subsection{Problem Setting}

We consider the viscous Burgers' equation as a canonical benchmark for developing and validating our defense methodology before scaling to the higher-dimensional CFD systems studied in~\citet{roy2026adversarial}:
\begin{equation}
    \frac{\partial u}{\partial t} + u \frac{\partial u}{\partial x} = \nu \frac{\partial^2 u}{\partial x^2}, \quad x \in [0, 2\pi], \quad t \in [0, T],
\end{equation}
with periodic boundary conditions and viscosity $\nu = 0.1$. The neural operator learns the solution map $\mathcal{G}^*: u_0 \mapsto u(\cdot, T)$ from initial conditions to solutions at terminal time $T=1$.

\paragraph{Adversarial perturbation model.} Following~\citet{roy2026adversarial}, we consider Gaussian perturbations applied in the discretized input space:
\begin{equation}
    \delta_j = m \cdot \exp\left(-\frac{(j - j_c)^2}{2\sigma^2}\right), \quad j = 0, 1, \ldots, n_x - 1,
\end{equation}
where $j_c$ is the center grid index, $m \in [-0.3, 0.3]$ is the magnitude, and $\sigma = 5$ (in grid-index units) controls the width. At this setting, the perturbation's $\pm 2\sigma$ range spans approximately 20 out of 64 grid points, affecting roughly 30\% of the discretized domain. This produces a semi-localized bump that models realistic scenarios such as sensor noise affecting a cluster of neighboring measurement points or calibration drift in a spatial region. For context, the maximum perturbation magnitude ($|m| = 0.3$) represents approximately 30--60\% of the typical initial condition peak amplitude (generated from 4-mode Fourier series with per-mode amplitudes of $0.25/k$), placing the threat model in a regime where the perturbation is physically significant but does not dominate the input signal.

\paragraph{Evaluation metrics.} We evaluate each strategy along two axes. The \textbf{baseline error} is the relative $L_2$ error on clean (unperturbed) test data, measuring standard prediction accuracy. The \textbf{robustness error} is the relative $L_2$ error between the model prediction under adversarial perturbation (found by DE) and the true physics solution of the perturbed input obtained from the numerical solver. This metric directly quantifies the model's deviation from physics under attack: a robust model should produce outputs close to the solver's response for the same perturbed input, since the well-posedness of the PDE guarantees that the true solution changes only slightly under small perturbations. The \textbf{combined score} is the sum of both errors, providing a single metric that captures the accuracy--robustness trade-off. A useful model must score low on both terms simultaneously.

\subsection{Active Learning for Robust Data Generation}

\begin{figure}[t]
\centering
\begin{tikzpicture}[
    node distance=0.6cm and 0.8cm,
    block/.style={rectangle, draw, rounded corners, minimum width=1.8cm, minimum height=0.7cm, align=center, font=\scriptsize},
    arrow/.style={-Stealth, thick},
    dashedarrow/.style={-Stealth, thick, dashed}
]
    \node[block, fill=primaryblue!20] (train) {Train\\Model};
    \node[block, fill=accentorange!20, right=of train] (probe) {Probe\\Weaknesses};
    \node[block, fill=purpleaccent!20, right=of probe] (eval) {Evaluate\\Both Metrics};
    \node[block, fill=dangerred!20, below=of eval] (adapt) {Adapt\\Smooth Ratio};
    \node[block, fill=successgreen!20, below=of probe] (gen) {Generate\\Targeted Data};
    
    \draw[arrow] (train) -- (probe);
    \draw[arrow] (probe) -- (eval);
    \draw[arrow] (eval) -- (adapt);
    \draw[arrow] (adapt) -- (gen);
    \draw[arrow] (gen) -| (train);
    
    \node[block, fill=lightgray, right=1.5cm of eval, minimum width=2cm] (de) {DE Attack:\\$\max_\delta \|f_\theta(u_0{+}\delta) - \mathcal{G}^*(u_0{+}\delta)\|$};
    \draw[dashedarrow] (probe) -- ++(0,1.0) -| (de);
    
    \node[above=1.1cm of probe, font=\scriptsize\itshape] {Discover vulnerabilities};
    \node[right=0.1cm of adapt, font=\scriptsize\itshape, align=left] {Baseline\\safeguard};
    
\end{tikzpicture}
\caption{Active learning loop with adaptive baseline safeguards. The current model is probed for weaknesses via differential evolution; targeted training data is then generated at vulnerability locations and paired with physics-corrected labels from the numerical solver. The smooth-ratio safeguard $\alpha$ prevents baseline degradation.}
\label{fig:active_learning}
\end{figure}
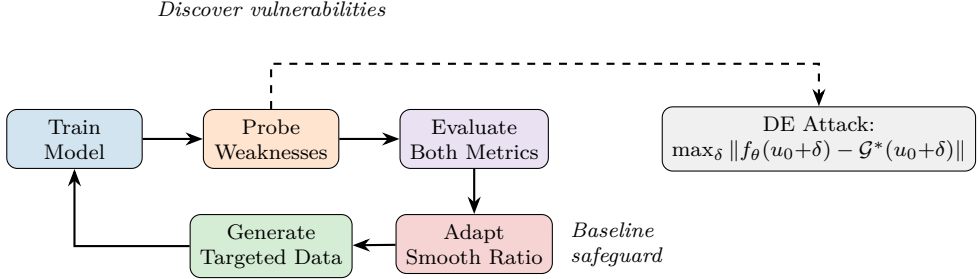

Our active learning strategy (Algorithm~\ref{alg:active_learning}, Figure~\ref{fig:active_learning}) operates in an iterative loop that couples adversarial probing with targeted data generation. At each round, the current model is first probed for weaknesses by running DE attacks on a held-out validation set. Both baseline and robustness metrics are then evaluated to track the accuracy--robustness trade-off, and the smooth-ratio safeguard $\alpha$ is adapted accordingly. New training data is generated targeting the discovered vulnerability locations, with physics labels obtained from the numerical solver, and the model is retrained on the augmented dataset.

\begin{algorithm}[t]
\caption{Active Learning with Adaptive Baseline Safeguards}
\label{alg:active_learning}
\begin{algorithmic}[1]
\REQUIRE Simulation budget $B$, bootstrap size $n_0$, samples per round $n_r$, baseline threshold $\tau$
\STATE Initialize: $\mathcal{D} \gets \text{SampleSmoothICs}(n_0)$, $\alpha \gets 0.4$
\STATE Train model $f_\theta$ on $\mathcal{D}$
\STATE $e_{\text{base}}^{\text{prev}} \gets \text{BaselineError}(f_\theta)$
\WHILE{$|\mathcal{D}| < B$}
    \STATE $\mathcal{W} \gets \text{ProbeWeaknesses}(f_\theta, \mathcal{D}_{\text{val}})$ \COMMENT{DE attacks on validation set}
    \STATE $e_{\text{base}}, e_{\text{robust}} \gets \text{Evaluate}(f_\theta)$
    \IF{$e_{\text{base}} > 1.1 \cdot e_{\text{base}}^{\text{prev}}$}
        \STATE $\alpha \gets \min(\alpha + 0.1, 0.7)$ \COMMENT{Increase smooth data fraction}
    \ELSIF{$e_{\text{base}} < 0.9 \cdot e_{\text{base}}^{\text{prev}}$ \textbf{and} $e_{\text{base}} < \tau$}
        \STATE $\alpha \gets \max(\alpha - 0.05, 0.3)$ \COMMENT{Allow more targeted data}
    \ENDIF
    \STATE $\mathcal{D}_{\text{new}} \gets \text{GenerateTargeted}(n_r, \mathcal{W}, \alpha)$
    \STATE $\mathcal{D} \gets \mathcal{D} \cup \mathcal{D}_{\text{new}}$
    \STATE Retrain $f_\theta$ on $\mathcal{D}$
    \STATE $e_{\text{base}}^{\text{prev}} \gets e_{\text{base}}$
\ENDWHILE
\RETURN $f_\theta$
\end{algorithmic}
\end{algorithm}

The central design element is the \textbf{adaptive smooth-ratio safeguard}. The parameter $\alpha \in [0.3, 0.7]$ controls the minimum fraction of smooth (unperturbed) samples \emph{within each round of $n_r$ new samples}: of the $n_r$ samples generated per round, at least $\lceil \alpha \cdot n_r \rceil$ are smooth ICs and the remainder target discovered weaknesses. If the baseline error increases beyond a 10\% relative tolerance, $\alpha$ is raised to restore accuracy; if baseline performance is satisfactory and below threshold $\tau$, $\alpha$ is lowered to allocate more of the fixed per-round budget toward targeted (perturbed) samples. This feedback mechanism prevents the common failure mode in adversarial training where robustness gains come at the expense of clean-data performance~\citep{tsipras2019robustness}.

A key distinction from classical adversarial training~\citep{madry2018towards} is that our targeted samples are paired with \emph{physics-corrected labels}, i.e., the true simulator output for the perturbed input, rather than with the unperturbed label used to penalize sensitivity. Classical adversarial training teaches a model to resist perturbations by maintaining its original prediction; our approach instead teaches the model to \emph{track the physics}, producing the correct solution for the perturbed input just as the solver would. This directly addresses the deviation-from-physics problem illustrated in Figure~\ref{fig:problem}, and is conceptually aligned with the multi-fidelity training (MFT) paradigm proposed in~\citet{roy2026adversarial}.

\subsection{Input Denoising Architecture}

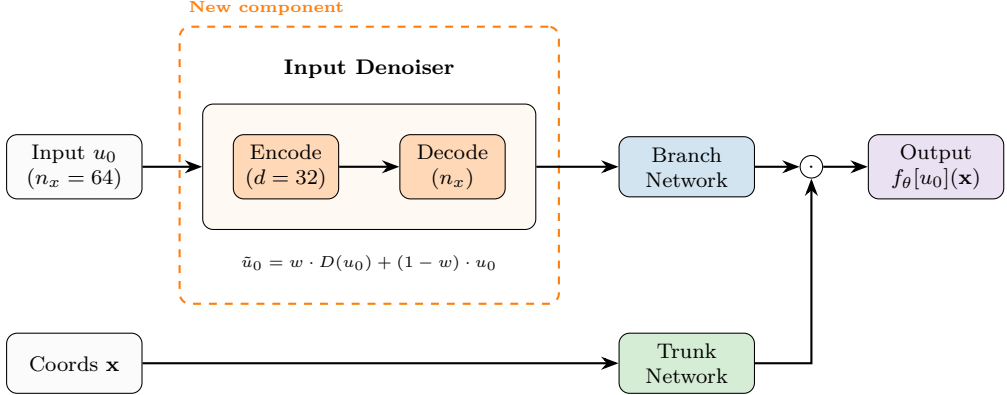
\begin{figure}[H]
\centering
\begin{tikzpicture}[
    node distance=1.2cm and 1.2cm,
    block/.style={rectangle, draw, rounded corners, minimum width=1.8cm, minimum height=0.8cm, align=center, font=\scriptsize},
    innerblock/.style={rectangle, draw, rounded corners, minimum width=1.4cm, minimum height=0.7cm, align=center, font=\scriptsize},
    arrow/.style={-Stealth, thick}
]
    \node[block, fill=lightgray!30] (input) {Input $u_0$\\$(n_x = 64)$};
    \node[innerblock, fill=accentorange!30, right=1.2cm of input] (enc) {Encode\\$(d=32)$};
    \node[innerblock, fill=accentorange!30, right=0.8cm of enc] (dec) {Decode\\$(n_x)$};
    \node[block, fill=primaryblue!20, right=1.5cm of dec] (branch) {Branch\\Network};
    \node[block, fill=purpleaccent!20, right=1.5cm of branch] (output) {Output\\$f_\theta[u_0](\mathbf{x})$};
    \node[block, fill=successgreen!20, below=1.8cm of branch] (trunk) {Trunk\\Network};
    \node[block, fill=lightgray!30] (coords) at (input |- trunk) {Coords $\mathbf{x}$};
    \begin{scope}[on background layer]
        \node[rectangle, draw, rounded corners, fill=accentorange!5, 
              inner sep=0.4cm, fit=(enc)(dec)] (denoise_box) {};
    \end{scope} 
    \node[above=0.2cm of denoise_box.north, font=\scriptsize\bfseries] (denoiser_label) {Input Denoiser};
    \node[below=0.2cm of denoise_box.south, font=\tiny] (residual_eq) {$\tilde{u}_0 = w \cdot D(u_0) + (1-w) \cdot u_0$};

    \node[draw, dashed, accentorange, thick, rounded corners, inner sep=0.3cm, 
          fit=(denoiser_label)(denoise_box)(residual_eq)] (highlight) {};
    \node[accentorange, font=\tiny\bfseries, anchor=south west] at (highlight.north west) {New component};
    \node[circle, draw, fill=white, inner sep=1.5pt, font=\scriptsize] at ($(branch.east)!0.5!(output.west)$) (dot) {$\cdot$};
    \draw[arrow] (input) -- (denoise_box);
    \draw[arrow] (enc) -- (dec);
    \draw[arrow] (denoise_box) -- (branch);
    \draw[arrow] (branch) -- (dot);
    \draw[arrow] (dot) -- (output);
    \draw[arrow] (coords) -- (trunk);
    \draw[arrow] (trunk) -| (dot);
\end{tikzpicture}
\caption{Input Denoising DeepONet architecture. A learnable autoencoder bottleneck ($64 \to 32 \to 64$) with a residual connection is placed before the branch network. The bottleneck attenuates adversarial perturbations while the learnable blend weight $w$ controls denoising intensity.}
\label{fig:architecture}
\end{figure}

We augment the standard DeepONet with a learnable input denoising layer placed before the branch network (Figure~\ref{fig:architecture}). The denoiser consists of a small autoencoder with a bottleneck:
\begin{equation}
    \tilde{u}_0 = w \cdot D(u_0) + (1-w) \cdot u_0,
\end{equation}
where $D: \mathbb{R}^{n_x} \to \mathbb{R}^{d} \to \mathbb{R}^{n_x}$ is an encoder--decoder pair with bottleneck dimension $d < n_x$, and $w = \mathrm{sigmoid}(\omega) \in (0,1)$ is a learnable blend weight controlled by a scalar parameter $\omega$.

The bottleneck enforces a compressed representation of the input function, which retains the dominant physics-relevant modes while attenuating high-frequency adversarial perturbations. The residual connection, modulated by the learnable weight $w$, allows the network to adaptively balance denoising strength against information preservation. During training, the entire system (denoiser + DeepONet) is optimized end-to-end, so the bottleneck learns to compress in a manner that is complementary to the downstream operator's reconstruction loss.

This design is motivated by the observation from~\citet{roy2026adversarial} that adversarial perturbations in neural operators tend to be structured and concentrated relative to the smooth physics solutions: a low-dimensional bottleneck is expected to attenuate these components preferentially.

\subsection{Combined Defense}
The combined approach trains an Input Denoising DeepONet using actively generated data. At the data level, active learning generates training pairs at vulnerability locations with physics-corrected labels, teaching the model the correct response to perturbation-like inputs. At the architecture level, input denoising reduces the effective input dimensionality via the bottleneck, attenuating adversarial components regardless of their spatial location. Neither defense is sufficient alone: active learning cannot anticipate all possible perturbation locations at test time, and input denoising cannot compensate for a model that has never learned the physics of perturbed inputs. The combination addresses both limitations.

\section{Experiments}

\subsection{Experimental Setup}

We use a spectral solver for the viscous Burgers' equation with $n_x = 64$ spatial discretization points, viscosity $\nu = 0.1$, and terminal time $T = 1.0$. All strategies are constrained to a total of 600 physics simulations, reflecting the computational cost typical of high-fidelity solvers in engineering applications. For the non-adaptive strategies (Baseline, Balanced, Denoising only), all 600 simulations are generated upfront; for the active learning strategies, the budget is dynamically allocated across iterations. We set bootstrap size $n_0 = 50$, samples per round $n_r = 20$, initial smooth ratio $\alpha_0 = 0.4$, and baseline threshold $\tau = 5\%$; the DE attack uses 30 iterations during probing and 40 during final evaluation. The base architecture is a standard DeepONet with 3-layer branch and trunk networks (hidden dimension 128, latent dimension 128), and the input denoising variant adds a bottleneck layer ($64 \to 32 \to 64$) with Tanh activations and a learnable blend weight. All models are trained with the Adam optimizer at an initial learning rate of $10^{-3}$ with cosine annealing, using 200 epochs for initial training and 100 epochs for active learning updates.

\subsection{Compared Strategies}

We compare five strategies spanning the data--architecture design space. The \textbf{Baseline} trains a standard DeepONet on smooth initial conditions only. The \textbf{Balanced} strategy uses the same architecture but trains on a mixture of 60\% perturbed and 40\% smooth ICs, where perturbations are random Gaussian bumps added to smooth profiles. The \textbf{Denoising only} strategy trains an Input Denoising DeepONet on smooth data. \textbf{Active Learning (AL)} applies the adaptive targeting procedure of Algorithm~\ref{alg:active_learning} with a standard DeepONet. Finally, \textbf{AL + Denoising} combines the adaptive targeting procedure with the Input Denoising DeepONet.

\subsection{Results}

\begin{table}[t]
\centering
\caption{Performance comparison across defense strategies. All methods operate under the same 600-simulation budget. Combined = Baseline + Robustness error. Lower is better for all metrics.}
\label{tab:results}
\begin{tabular}{@{}lccccc@{}}
\toprule
\textbf{Strategy} & \textbf{Data} & \textbf{Model} & \textbf{Baseline (\%)} & \textbf{Robust (\%)} & \textbf{Combined (\%)} \\
\midrule
Baseline & Smooth & Standard & 3.27 & 12.15 & 15.42 \\
Balanced & Mixed & Standard & 3.97 & 4.53 & 8.49 \\
Denoising only & Smooth & Denoising & 2.59 & 2.64 & 5.22 \\
Active Learning & Adaptive & Standard & 1.69 & 1.74 & 3.42 \\
\rowcolor{boxgreen}
\textbf{AL + Denoising} & \textbf{Adaptive} & \textbf{Denoising} & \textbf{1.21} & \textbf{0.83} & \textbf{2.04} \\
\bottomrule
\end{tabular}
\end{table}

\begin{figure}[t]
\centering
\begin{tikzpicture}
\begin{axis}[
    width=0.9\linewidth,
    height=6cm,
    ybar,
    bar width=0.35cm,
    ylabel={Error (\%)},
    symbolic x coords={Baseline, Balanced, Denoising only, Active Learning, AL + Denoising},
    xtick=data,
    x tick label style={rotate=20, anchor=east, font=\small},
    ymin=0,
    ymax=14,
    legend style={at={(0.98,0.98)}, anchor=north east, font=\small},
    legend cell align={left},
    grid=major,
    grid style={dashed, gray!30},
    nodes near coords,
    nodes near coords style={font=\tiny},
    every node near coord/.append style={anchor=south},
]
\addplot[fill=primaryblue!70] coordinates {
    (Baseline, 3.27)
    (Balanced, 3.97)
    (Denoising only, 2.59)
    (Active Learning, 1.69)
    (AL + Denoising, 1.21)
};
\addplot[fill=dangerred!70] coordinates {
    (Baseline, 12.15)
    (Balanced, 4.53)
    (Denoising only, 2.64)
    (Active Learning, 1.74)
    (AL + Denoising, 0.83)
};
\legend{Baseline Error, Robustness Error}
\end{axis}
\end{tikzpicture}
\caption{Baseline and robustness errors across defense strategies. The combined approach (AL + Denoising) achieves the lowest error on both metrics, demonstrating that data-level and architecture-level defenses provide complementary gains.}
\label{fig:results_bar}
\end{figure}
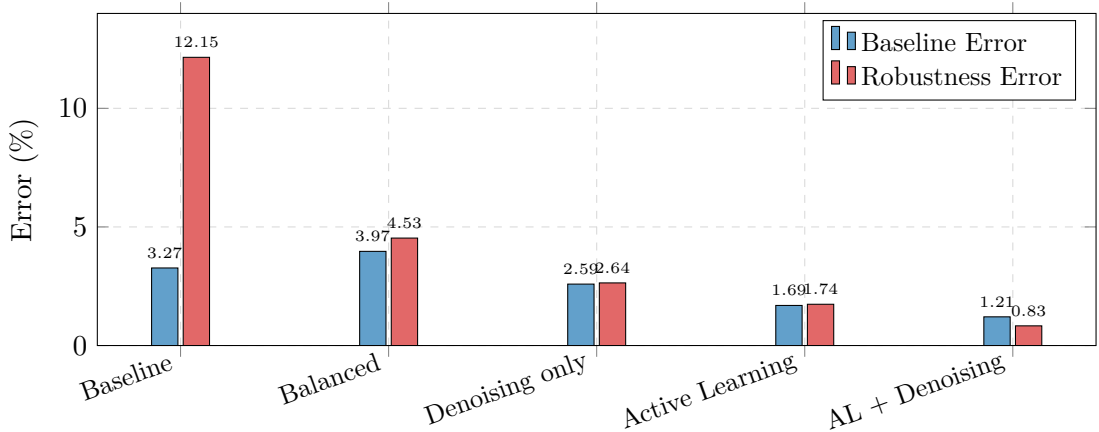

Table~\ref{tab:results} and Figure~\ref{fig:results_bar} present the main results. The baseline strategy achieves 3.27\% error on clean data but 12.15\% under adversarial attack, approximately a $3.7\times$ degradation. This confirms that validation accuracy alone is not predictive of adversarial performance, consistent with the broader findings of~\citet{roy2026adversarial}. The balanced strategy reduces robustness error to 4.53\% (a 63\% improvement) by exposing the model to perturbed inputs during training, though this comes at a modest cost to baseline performance (3.97\% vs.\ 3.27\%), illustrating the accuracy--robustness trade-off observed in adversarial training~\citep{tsipras2019robustness}.

The denoising architecture trained on smooth data alone reaches 2.64\% robustness error (5.22\% combined), outperforming the naive balanced strategy and demonstrating that architectural modifications can provide meaningful robustness even without targeted data. However, it falls short of the next strategy. Active learning achieves substantially stronger results on both metrics: by targeting discovered weaknesses, it reaches 1.74\% robustness error (86\% improvement over baseline) while \emph{simultaneously improving} baseline accuracy to 1.69\%. This dual improvement is attributable to two factors: targeted data is more informative per sample than random sampling, and the adaptive smooth-ratio safeguard explicitly prevents baseline degradation.

Adding input denoising to active learning further reduces robustness error to 0.83\% (a 52\% improvement over AL alone) and baseline error to 1.21\%, yielding a combined score of 2.04\%, an \textbf{87\% reduction} relative to standard training. The combined approach Pareto-dominates all other strategies on both metrics.

\subsection{Ablation: Why the Combination Works}

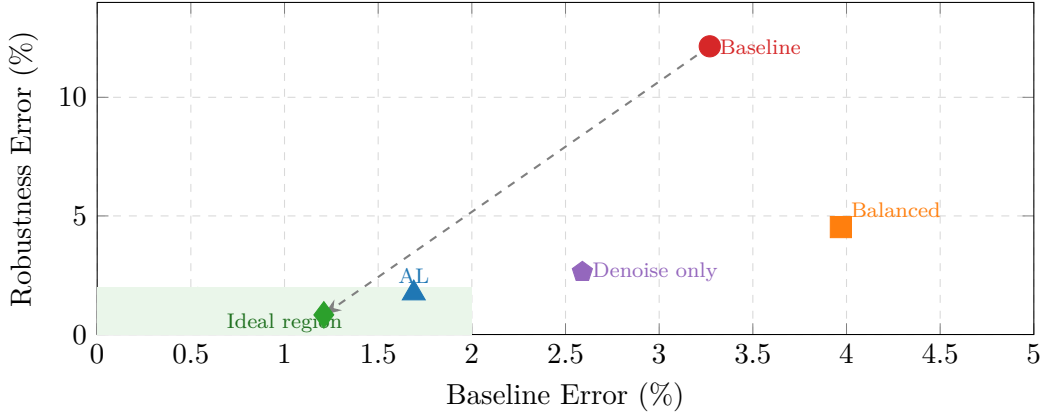
\begin{figure}[t]
\centering
\begin{tikzpicture}
\begin{axis}[
    width=0.85\linewidth,
    height=6cm,
    xlabel={Baseline Error (\%)},
    ylabel={Robustness Error (\%)},
    xmin=0, xmax=5,
    ymin=0, ymax=14,
    grid=major,
    grid style={dashed, gray!30},
]
\addplot[only marks, mark=*, mark size=4pt, dangerred] 
    coordinates {(3.27, 12.15)} node[right, font=\scriptsize] {Baseline};
\addplot[only marks, mark=square*, mark size=4pt, accentorange] 
    coordinates {(3.97, 4.53)} node[above right, font=\scriptsize] {Balanced};
\addplot[only marks, mark=triangle*, mark size=5pt, primaryblue] 
    coordinates {(1.69, 1.74)} node[above, font=\scriptsize] {AL};
\addplot[only marks, mark=diamond*, mark size=5pt, successgreen] 
    coordinates {(1.21, 0.83)} node[above left, font=\scriptsize] {AL+Denoise};
\addplot[only marks, mark=pentagon*, mark size=4pt, purpleaccent] 
    coordinates {(2.59, 2.64)} node[right, font=\scriptsize] {Denoise only};

\fill[successgreen!10] (0,0) rectangle (2,2);
\node[font=\scriptsize, successgreen!70!black] at (1,0.5) {Ideal region};

\draw[-Stealth, thick, gray, dashed] (3.27, 12.15) -- (1.21, 0.83);

\end{axis}
\end{tikzpicture}
\caption{Accuracy--robustness trade-off across defense strategies. The green shaded region marks the ideal low-error corner. The combined approach (green diamond) Pareto-dominates all alternatives, including input denoising alone (purple pentagon) and active learning alone (blue triangle). The dashed arrow indicates the improvement trajectory from standard training.}
\label{fig:ablation}
\end{figure}

To understand why the combination outperforms individual components, Figure~\ref{fig:ablation} maps all five configurations in the baseline--robustness error plane. The ablation reveals that neither component alone reaches the ideal low-error corner. Input denoising without active learning fails to achieve optimal robustness because the denoiser cannot fully compensate for a model that has not learned the physics of perturbed inputs; the bottleneck filters noise but cannot correct the operator's response to input patterns outside its training distribution. Conversely, active learning without denoising achieves strong robustness but remains somewhat sensitive to perturbations outside the specific locations probed during training. The combination addresses both failure modes: active learning provides coverage of the perturbation landscape, while the denoiser provides a continuous defense that generalizes beyond the training distribution.

\section{Discussion}

\subsection{Relationship to Prior Work}

Our defense framework builds directly on the vulnerability analysis of~\citet{roy2026adversarial}, which characterized the sensitivity mismatch problem and introduced the $d_{\text{eff}}$ diagnostic for neural operators deployed in nuclear digital twin applications~\citep{kobayashi2024deeponet,hossain2024virtual}. While that work focused on attack characterization, we demonstrate here that the identified mismatch can be partially mitigated through complementary data- and architecture-level interventions.

The active learning component shares conceptual ground with adversarial training~\citep{madry2018towards} and multi-fidelity training~\citep{roy2026adversarial}, but with an important distinction: we train on \emph{physics-corrected} responses rather than adversarial labels, teaching the model the true input--output mapping for perturbed conditions. This avoids the accuracy--robustness tension that plagues standard adversarial training~\citep{tsipras2019robustness} and aligns with the MFT philosophy that simulator data is the authoritative source of robustness.

The input denoising component relates to purification-based defenses in image classification~\citep{shi2021online} and defensive distillation~\citep{papernot2016distillation}, but is adapted to the operator learning setting where adversarial perturbations are structured and semi-localized rather than globally distributed. The learnable blend weight provides a principled mechanism for balancing noise removal against information loss, a trade-off that is especially important when the input function contains sharp physical features that might otherwise be misidentified as adversarial perturbations.

\subsection{Hypothesis: Optimal Training Data is Architecture-Dependent}

A broader hypothesis motivated by our active learning framework, when considered alongside the cross-architecture vulnerability analysis of~\citet{roy2026adversarial}, is that \emph{uniform data sampling is a suboptimal training strategy for neural operators}, and the degree of suboptimality may vary substantially across architectures. This hypothesis follows from a simple but consequential chain of reasoning.

The Jacobian analysis in~\citet{roy2026adversarial}, conducted on a 102-dimensional heat exchanger CFD benchmark, revealed that each architecture concentrates its input sensitivity in a distinct subspace. S-DeepONet's GRU-based encoder creates extreme sensitivity at sequence endpoints, with a near-perfect correlation ($r = 0.99$) between Jacobian column norms and DE attack targeting frequency. POD-DeepONet concentrates virtually all sensitivity in the global inlet parameters, with boundary condition Jacobian norms twelve orders of magnitude smaller and attack targeting that is effectively random ($r = -0.47$). MIMONet and NOMAD distribute sensitivity more broadly ($d_{\text{eff}} \approx 32$), with moderate targeting correlations ($r = 0.50$ and $0.57$ respectively).

If one were to run our active learning procedure separately for each architecture, the DE probing phase would discover fundamentally different vulnerability locations, and the subsequent data generation phase would produce fundamentally different training sets. S-DeepONet's active learner would concentrate perturbations at sequence boundaries; POD-DeepONet's would focus on global parameter variations; MIMONet and NOMAD would require broader input-space coverage. A single uniformly sampled training set cannot simultaneously address all of these architecture-specific vulnerability patterns.

This has a practical consequence: the common practice of training multiple neural operator architectures on the same dataset and selecting the best performer conflates two distinct sources of error. A model may underperform not because its architecture is inferior, but because the training data does not adequately cover its specific vulnerability subspace. Architecture-aware data generation, as provided by our active learning framework, decouples these factors and may reveal different performance rankings than uniform-data comparisons would suggest. While we demonstrate this principle on a single architecture in the present work, extending the active learning procedure across architectures on the CFD benchmark of~\citet{roy2026adversarial} is a natural next step.

\subsection{Limitations and Future Directions}

These results are preliminary, established on a single 1D benchmark (Burgers' equation), and several extensions are necessary before deployment-ready conclusions can be drawn. The current evaluation uses the same attack family (DE) for both active learning probing and final robustness assessment; validating against unseen attack types such as PGD or transfer attacks would strengthen the generalization claim. All results are reported from single runs; future work should include error bars over multiple random seeds to confirm statistical significance of the inter-strategy differences. The combined score weights baseline and robustness errors equally, which is a simplifying choice; in practice, the relative importance of clean-data accuracy versus adversarial robustness will depend on the deployment context. A comparison against classical Madry-style adversarial training (using worst-case perturbations with original labels under the same simulation budget) would further contextualize our physics-corrected approach.

On the benchmarking side, validation on 2D Navier--Stokes, Darcy flow, and coupled thermal-hydraulic systems is needed to confirm generalizability; the heat exchanger CFD benchmark of~\citet{roy2026adversarial}, with its 102-dimensional input space and multi-channel field outputs, would be a natural next target. The current results are also limited to DeepONet, and testing on architectures with fundamentally different sensitivity profiles, particularly FNO~\citep{kovachki2023neural} and the multi-output variants examined in~\citet{roy2026adversarial}, is essential.

On the methodological side, the iterative probe--generate--retrain loop adds computational overhead that may become prohibitive for high-fidelity 3D simulations; budget-efficient variants such as surrogate-assisted DE or transfer of vulnerability maps across architectures deserve investigation. A formal theoretical characterization of when and why the data--architecture combination provides synergistic gains, potentially through the lens of $d_{\text{eff}}$ reduction, remains an open problem. Finally, integrating certified robustness mechanisms such as randomized smoothing~\citep{cohen2019certified} or Lipschitz-constrained architectures~\citep{gouk2021regularisation} could provide provable robustness bounds for neural operators, moving beyond the empirical guarantees presented here.

\subsection{Implications for Deployment}

For practitioners deploying neural operators in safety-critical energy systems such as nuclear digital twins~\citep{kobayashi2024deeponet,hossain2024virtual} or real-time virtual sensors~\citep{kobayashi2025conformalized}, our results carry several practical implications. Standard training, regardless of how low the validation error appears, is insufficient for adversarial robustness. Data augmentation with perturbed inputs provides meaningful improvement, but adaptive targeting through active learning is substantially more sample-efficient. Architectural modifications that reduce input sensitivity complement data-level defenses. Perhaps most importantly, robustness must be evaluated explicitly; validation accuracy is not a reliable proxy for adversarial performance.

\section{Conclusion}

We have presented preliminary results demonstrating that active learning and input denoising provide synergistic defense against adversarial attacks on neural operators. On the viscous Burgers' equation benchmark, the combined approach achieves a 2.04\% combined error, an 87\% reduction relative to standard training, by addressing complementary aspects of the vulnerability: active learning teaches the model \emph{where} perturbations occur and how to respond correctly, while input denoising provides \emph{inherent} architectural robustness via dimensionality reduction of the input representation.

These results establish a promising foundation for developing robust neural operator surrogates for safety-critical energy systems. Beyond the defense itself, our findings, together with prior cross-architecture vulnerability analysis~\citep{roy2026adversarial}, suggest the hypothesis that optimal training data for neural operators is architecture-dependent: because different architectures concentrate sensitivity in different input subspaces, a single uniformly sampled dataset cannot adequately cover the vulnerability landscape of all models. Ongoing work extends validation to multi-dimensional PDE benchmarks and real-world nuclear thermal-hydraulic data~\citep{roy2026adversarial,kobayashi2024deeponet}, with the goal of establishing deployment-ready robustness assurances for digital twin systems.

\section*{Acknowledgments}
This work used the Delta and DeltaAI systems at the National Center for Supercomputing Applications [awards OAC 2005572 and OAC 2320345] through allocation CIS240093 from the Advanced Cyberinfrastructure Coordination Ecosystem: Services \& Support (ACCESS) program, which is supported by National Science Foundation grants \#2138259, \#2138286, \#2138307, \#2137603, and \#2138296.

\bibliographystyle{plainnat}

\begin{thebibliography}{20}

\bibitem[Adesoji and Chen(2022)]{adesoji2022evaluating}
Adesoji, A.~D. and Chen, P.-Y.
\newblock Evaluating the adversarial robustness for {F}ourier neural operators.
\newblock \emph{arXiv preprint arXiv:2204.04259}, 2022.

\bibitem[Cohen et al.(2019)]{cohen2019certified}
Cohen, J., Rosenfeld, E., and Kolter, Z.
\newblock Certified adversarial robustness via randomized smoothing.
\newblock In \emph{International Conference on Machine Learning}, pages 1310--1320, 2019.

\bibitem[Gouk et al.(2021)]{gouk2021regularisation}
Gouk, H., Frank, E., Pfahringer, B., and Cree, M.~J.
\newblock Regularisation of neural networks by enforcing {L}ipschitz continuity.
\newblock \emph{Machine Learning}, 110(2):393--416, 2021.

\bibitem[Hossain et al.(2024)]{hossain2024virtual}
Hossain, R.~B., Ahmed, F., Kobayashi, K., Koric, S., Abueidda, D., and Alam, S.~B.
\newblock Virtual sensing-enabled digital twin framework for real-time monitoring of nuclear systems leveraging deep neural operators.
\newblock \emph{arXiv preprint arXiv:2410.13762}, 2024.

\bibitem[Kobayashi and Alam(2024)]{kobayashi2024deeponet}
Kobayashi, K. and Alam, S.~B.
\newblock Deep neural operator-driven real-time inference to enable digital twin solutions for nuclear energy systems.
\newblock \emph{Scientific Reports}, 14:3935, 2024.

\bibitem[Kobayashi et al.(2024)]{kobayashi2024generalization}
Kobayashi, K., Daniell, J., and Alam, S.~B.
\newblock Improved generalization with deep neural operators for engineering systems: Path towards digital twin.
\newblock \emph{Engineering Applications of Artificial Intelligence}, 131:107844, 2024.

\bibitem[Kobayashi et al.(2025a)]{kobayashi2025proxies}
Kobayashi, K., Roy, S., Koric, S., Abueidda, D., and Alam, S.~B.
\newblock From proxies to fields: Spatiotemporal reconstruction of global radiation from sparse sensor sequences.
\newblock \emph{arXiv preprint arXiv:2506.12045}, 2025.

\bibitem[Kobayashi et al.(2025b)]{kobayashi2025conformalized}
Kobayashi, K., Garg, S., Ahmed, F., Chakraborty, S., and Alam, S.~B.
\newblock Distribution-free uncertainty-aware virtual sensing via conformalized neural operators.
\newblock \emph{arXiv preprint arXiv:2507.11574}, 2025.

\bibitem[Kovachki et al.(2023)]{kovachki2023neural}
Kovachki, N., Li, Z., Liu, B., Azizzadenesheli, K., Bhattacharya, K., Stuart, A., and Anandkumar, A.
\newblock Neural operator: Learning maps between function spaces with applications to {PDE}s.
\newblock \emph{Journal of Machine Learning Research}, 24(89):1--97, 2023.

\bibitem[Lu et al.(2021)]{lu2021learning}
Lu, L., Jin, P., Pang, G., Zhang, Z., and Karniadakis, G.~E.
\newblock Learning nonlinear operators via {DeepONet} based on the universal approximation theorem of operators.
\newblock \emph{Nature Machine Intelligence}, 3(3):218--229, 2021.

\bibitem[Lu et al.(2022)]{lu2022comprehensive}
Lu, L., Meng, X., Cai, S., Mao, Z., Goswami, S., Zhang, Z., and Karniadakis, G.~E.
\newblock A comprehensive and fair comparison of two neural operators (with practical extensions) based on {FAIR} data.
\newblock \emph{Computer Methods in Applied Mechanics and Engineering}, 393:114778, 2022.

\bibitem[Madry et al.(2018)]{madry2018towards}
Madry, A., Makelov, A., Schmidt, L., Tsipras, D., and Vladu, A.
\newblock Towards deep learning models resistant to adversarial attacks.
\newblock In \emph{International Conference on Learning Representations}, 2018.

\bibitem[Miyato et al.(2018)]{miyato2018spectral}
Miyato, T., Kataoka, T., Koyama, M., and Yoshida, Y.
\newblock Spectral normalization for generative adversarial networks.
\newblock In \emph{International Conference on Learning Representations}, 2018.

\bibitem[Papernot et al.(2016)]{papernot2016distillation}
Papernot, N., McDaniel, P., Wu, X., Jha, S., and Swami, A.
\newblock Distillation as a defense to adversarial perturbations against deep neural networks.
\newblock In \emph{IEEE Symposium on Security and Privacy}, pages 582--597, 2016.

\bibitem[Roy et al.(2026)]{roy2026adversarial}
Roy, S., Kobayashi, K., Chakraborty, S., Rizwan-uddin, and Alam, S.~B.
\newblock Adversarial vulnerabilities in neural operator digital twins: Gradient-free attacks on nuclear thermal-hydraulic surrogates.
\newblock \emph{arXiv preprint arXiv:2603.22525}, 2026.

\bibitem[Shi et al.(2021)]{shi2021online}
Shi, C., Holtz, C., and Mishne, G.
\newblock Online adversarial purification based on self-supervised learning.
\newblock \emph{arXiv preprint arXiv:2101.09387}, 2021.

\bibitem[Tsipras et al.(2019)]{tsipras2019robustness}
Tsipras, D., Santurkar, S., Engstrom, L., Turner, A., and Madry, A.
\newblock Robustness may be at odds with accuracy.
\newblock In \emph{International Conference on Learning Representations}, 2019.

\bibitem[Zhu et al.(2023)]{zhu2023fourier}
Zhu, M., Feng, S., Lin, Y., and Lu, L.
\newblock Fourier-{DeepONet}: Fourier-enhanced deep operator networks for full waveform inversion with improved accuracy, generalizability, and robustness.
\newblock \emph{Computer Methods in Applied Mechanics and Engineering}, 416:116300, 2023.

\end{thebibliography}

\end{document}